\documentclass[preprint]{elsarticle}

\usepackage{lineno,hyperref}
\nolinenumbers

\usepackage[utf8]{inputenc} % allow utf-8 input
\usepackage[T1]{fontenc}    % use 8-bit T1 fonts
\usepackage{hyperref}       % hyperlinks
\usepackage{url}            % simple URL typesetting
\usepackage{booktabs}       % professional-quality tables
\usepackage{amsfonts}       % blackboard math symbols
\usepackage{nicefrac}       % compact symbols for 1/2, etc.
\usepackage{microtype}      % microtypography
\usepackage{xcolor}         % colors
\usepackage{graphicx}
\usepackage{amsfonts}
\usepackage[Algorithmus]{algorithm}
\usepackage{algorithmicx}
\usepackage{amsmath}
\usepackage{newtxmath}
\newcommand{\maxf}[1]{\ensuremath{\boldmath #1}}
\DeclareMathOperator*{\argmax}{arg\,max}

\usepackage{caption}
\usepackage{subcaption}
\usepackage{siunitx}
\usepackage{enumitem}

\usepackage{setspace}
\usepackage{algpseudocode}
\algnewcommand{\IIf}[1]{\State\algorithmicif\ #1\ \algorithmicthen}
\algnewcommand{\EndIIf}{\unskip\ \algorithmicend\ \algorithmicif}

\journal{Journal of \LaTeX\ Templates}

\begin{document}

\begin{frontmatter}

\title{Fast online inference for nonlinear contextual bandit based on Generative Adversarial Network}

\author{Yun Da Tsai, Shoe De Lin}
\address{bb04902103@gmail.com, sdlin@csie.ntu.edu.tw}
\address{Department of Computer Science, National Taiwan University\\
No. 1, Sec. 4, Roosevelt Rd., Taipei 10617, Taiwan}

% %% or include affiliations in footnotes:
% \author[mymainaddress,mysecondaryaddress]{Elsevier Inc}

% \author[mysecondaryaddress]{Global Customer Service\corref{mycorrespondingauthor}}
% \cortext[mycorrespondingauthor]{Corresponding author}

% \address[mymainaddress]{1600 John F Kennedy Boulevard, Philadelphia}
% \address[mysecondaryaddress]{360 Park Avenue South, New York}

\begin{abstract}
This work addresses the efficiency concern on inferring a nonlinear contextual bandit when the number of arms $n$ is very large.
  We propose a neural bandit model with an end-to-end training process to efficiently perform bandit algorithms such as Thompson Sampling and UCB during inference.
  We advance state-of-the-art time complexity to $O(\log n)$ with approximate Bayesian inference, neural random feature mapping, approximate global maxima and approximate nearest neighbor search.
  We further propose a generative adversarial network to shift the bottleneck of maximizing the objective for selecting optimal arms from inference time to training time, enjoying significant speedup with additional advantage of enabling batch and parallel processing. %The generative model can inference an approximate argmax of the posterior sampling in logarithmic time complexity with the help of approximate nearest neighbor search.
  Extensive experiments on classification and recommendation tasks demonstrate order-of-magnitude improvement in inference time no significant degradation on the performance.
\end{abstract}

\begin{keyword}
Neural Networks, Bandit, GAN, Thompson Sampling, UCB algorithm
\MSC[2010] 00-01\sep  99-00
\end{keyword}

\end{frontmatter}

% \linenumbers

\section{Introduction}
\label{sec:intro}
%intro - GLB
% In the stochastic contextual bandit problem, an agent is given a set of arms $A\inR^d$, each of which is associated with a context feature vector $x\in R^d$ at each time $t$.
% The agent then chooses an arm and receives a stochastic reward from some unknown distribution conditioned on the context.
% The goal is to maximize the cumulative reward over time.

Bandit algorithms for exploration-exploitation have attracted attentions from both academic and industrial communities.
The bandit agent learns in a stochastic environment to estimate the reward of each arm.
The goal is to eventually minimize regret that measures how much cumulative reward an agent gains by selecting different arms over a period of time.
Bandit algorithms have wide applications, for instance, online advertisements, recommendation ~\cite{toth2020balancing,guo2020deep} , information retrieval~\cite{kveton2015cascading,combes2015learning}, routing and network optimization~\cite{gyorgy2007line,gai2012combinatorial}, and influence maximization in social networks ~\cite{carpentier2016revealing}.

% motivation - scalability
Online recommendation applications normally require real-time responses given the context information.
Most existing bandit algorithms, such as Upper Confidence Bound (UCB) and Thompson Sampling (TS), find the optimal arm by maximizing the objective in every arm selection process, which prevents efficient online inference when the number of arms grows.
One can rely on linear models to ease such burden with asymptotically optimal algorithms, e.g. linear, Lipschitz and unimodal models~\cite{combes2017minimal}.
% e.g. linear~\cite{lattimore2017end}, Lipschitz~\cite{magureanu2014lipschitz} and unimodal~\cite{combes2014unimodal}.%combes2017minimal
Unfortunately, such trick does not work for nonlinear contextual bandits (also known as generalized linear bandits) and thus leads to the time complexity scales linearly with the number of arms $N$.
This would be impractical for applications seeking real-time response with a large number of arms, in particular when the arms are given by all points in a continuous set of dimensions $d$.%~\cite{agrawal2013thompson}.
In reality, a recommendation system generally has a large number of items as arms; or a retrieval system treats each document as a single arm. Near real-time response is required for both tasks in E-Commerce. 
% why neural
%In this work, we focus mainly on Neural Bandit models since for nonlinear contextual bandits, neural networks is especially appealing for its rich expressive power that no assumption is made about the reward function combining with efficient exploration mechanisms~\cite{phan2019thompson}.

% research problem
We would like to address the scalability issue in real-time online service for nonlinear neural contextual bandit models. Furthermore, we focus mainly on Neural Bandit models. For nonlinear contextual bandits, neural networks are especially appealing for its rich expressive power with no assumption made about the reward function combining with efficient exploration mechanisms~\cite{phan2019thompson}.
%We focus on accelerating the run-time at inference phase while doing online serving.
% approach
% Towards reducing the time complexity for arm selecting in every time step (requests), we proposed a scalable algorithm that integrates several approximation approaches which is performed end-to-end on neural networks and successfully transform the problem into fast approximate nearest neighbor search.
% This achieves a logarithmic time complexity although accompanied with several drawbacks such as a nontrivial overhead with large constant factor and likely to consume more memory if processed in batch or parallel.
% We further proposed GANBandit which is based on generative adversarial network that moves everything but necessary ahead of time from inference phase to training phase thus significantly improves the time and space complexity at the time of online serving with no drawbacks.
To achieve this goal, we first propose a scalable algorithm to perform end-to-end training and execution on neural networks by transforming the problem into a fast approximate nearest neighbor search problem.
This algorithm enjoys logarithmic time complexity for selecting the best arm in every single request.
To further improve the above framework, we then propose the GANBandit algorithm which is based on generative adversarial network to shift the costly computation burden from inference phase to training phase, further improve the time and space complexity by reducing the nontrivial overhead of back-propagation. It also allows the model to be efficiently processed in batch or parallel with only constant space complexity.
%Finally, through experiments we demonstrate that the proposed method could be applied to the continuum-arm bandit problem in generic metric space with infinite arms.

% comparison
%There are extensive works on the bandit problem in the literature and several previous works address different aspect of scalability issues in bandit algorithm which we will discuss more in section~\ref{sec:related}.
Experiments on both artificial and real datasets demonstrate that the proposed model can significantly reduce the inference time without apparent sacrifices on performance compared to with the conventional contextual bandit algorithms. 
%including linear bandits, nonlinear bandits with exhaust search through $N$ arms, the proposed scalable algorithm \textit{Multi} and \textit{GANBandit}.
We further conduct an experiment in a continuum-arm setting, and observe that the proposed solution obtains favorable results compared to GP-UCB~\cite{srinivas2009gaussian} and Hierarchical optimistic optimization (HOO)~\cite{bubeck2011x} algorithm in Appendix~\ref{app:continuum}.
%The experiment results demonstrate a competitive regret bound with significant speedup by many orders of magnitude.
% contribution
To our knowledge, this is the first work to achieve logarithmic time complexity and constant space complexity for inference in nonlinear contextual bandit.

\section{Related Works}
\label{sec:related}
\subsection{Contextual Bandit}
The most studied bandit model in the literature is linear contextual bandits~\cite{chu2011contextual,dani2008stochastic}.
Alongside there are many existing structures investigated, including: linear, combinatorial, Lipschitz, and unimodal bandits~\cite{combes2017minimal}.
There are also settings with infinitely many arms~\cite{abbasi2011improved,wang2008infinitely} and in generic continuous metric space~\cite{kleinberg2013bandits,bubeck2011x} with hierarchical tree-based space partition algorithms.
To deal with nonlinearity, generalized linear bandits have been considered.
GLM-UCB~\cite{filippi2010parametric} assumes that the reward function can be written as a composition of a linear function and a link function.
Others explore more general nonlinear bandits without making strong modeling assumptions. % and SupCB-GLM~\cite{li2017provably}
GP-UCB~\cite{srinivas2009gaussian} assumes that the reward function is generated from a Gaussian process with known mean and covariance functions. % and CGP-UCB~\cite{krause2011contextual}
KernelUCB~\cite{valko2013finite} assumes that the reward function lies in a RKHS with bounded RKHS norm.
Nevertheless, these methods require fairly strong assumptions on the reward function.

\subsection{Neural Bandit}
Recent advances in deep learning literature has helped researchers gain more understanding about neural networks in Bayesian settings which is adopted in bandit problems.
NeuralBandit~\cite{allesiardo2014neural} uses bootstrapping which consists of K neural networks.
\cite{urteaga2018variational} proposes variational inference in Thompson Sampling for contextual bandit.
\cite{lipton2018bbq,azizzadenesheli2018efficient} also use variational Thompson Sampling in reinforcement learning with deep-Q learning.
NeuralLinear~\cite{riquelme2018deep,zahavy2019deep} uses the former layers of neural networks as a feature map to transform contexts from raw input space to a better representation in low-dimensional space, then applies Thompson Sampling on the last layer to choose an action.
NeuralUCB~\cite{zhou2020neural} uses random feature mapping defined by the neural network gradient to construct the upper confidence bound for contextual bandit and provide a theoretical guarantee on the regret.

\subsection{Scalability Issues in Bandit}
\label{sec:scalability}
There are several different aspects to the scalability issues in bandit problems.
The scaling MAB problem proposed in~\cite{fouche2019scaling} focuses on the situation where evaluating arms could be costly such that the fewer arms evaluated the better.
They solve the bandit problem which maximizes cumulative reward under an efficiency constraint to reduce the number of arms played and minimizes the cost while keeping the regrets low.
GLOC~\cite{jun2017scalable} focuses on the scalability problem where the time step $T$ (or rounds) is very large.
Existing nonlinear bandit algorithm requires storing all the arms and rewards appeared so far as $a_{1:t-1}, x_{1:t-1}, r_{1:t-1}$.
The space complexity as well as the time complexity for batch optimization grows linearly with $T$.
The solution takes an online learning (OL) algorithm and transforms it into a bandit algorithm with a low regret bound with the help of a novel generalization online-to-confidence-set conversion technique.
Volumetric spanners~\cite{hazan2016volumetric} and QGLOC~\cite{jun2017scalable} address the challenge that is more similar to ours, focusing on the scalability issue where the number of arm sets is very large.
The former provides a simple approach to select a subset of arms ahead of time. This solution is specialized for efficient exploration only and may inadvertently rule out a large number of good arms.
The later transforms the maximizing objective into quadratic form which then can be solved by using approximate maximum inner product search hashing.
However, QGLOC requires the objective
function to be a distance or an inner product computation which can be satisfied by only a subset of models.
%After the hash table construction with Gaussian projection vectors, the hashing method achieves an online compute time complexity sublinear to $N$ and a nontrivial overhead of computing the projections to determine the hash keys while the proposed approach has a time complexity logarithmic to $N$.

\section{Preliminaries and the Basic Model}
\label{sec:basic}
\subsection{Contextual Bandit}
% set action a, time t, context x, reward r, distribution f
In this paper, we consider the structured stochastic contextual bandit problem and focus on a finite but very large number of arms.
Nevertheless, the proposed method is applicable in continuum-arm setting with infinite arms, which we will demonstrate in Appendix~\ref{app:continuum}.
Given a context vector $x_t\in R^d$ at time $t$, each action $a\in A := \{1,\cdots,N\} \in R^d$ could receive a reward $r_{a,x,t}$ drawn from an unknown distribution $f_a(r|x,\theta)$ parameterized by $\theta$.
% set history
We denote the history of given contexts, chosen arms, and observed rewards up to time $t$ as $x_{1:t-1} \equiv (x_1, \cdots , x_{t-1})$, $a_{1:t-1} \equiv (a_1, \cdots , a_{t-1})$ and $y_{1:t} \equiv (y_1, \cdots , y_{t-1})$, respectively.
% independent of history
The observed reward $r_t$ is independent of the history and is drawn from the reward distribution conditional to the chosen arm $a_t$, given context $x_t$ and $\theta$; i.e., $r_t \sim f_a(y|x_t, \theta)$.
% bandit algo
The bandit algorithm learns the reward distribution through interaction with the world by taking actions sequentially based on past history and given context.
The goal of the algorithm is to maximize the expected (cumulative) reward which is equivalent to minimize regrets.
We denote the optimal action at time $t$ and the regret as,
\begin{align}
    a_t^* := \argmax_{a  \in A}{E(r_{a,t}|x_{a,t})}\label{eq:argmax}, &~~Regret(t) = \sum_{t=1}^{T} r_{a*,t} - r_t
% \label{eq:regret}
\end{align}
.

\subsection{Thompson Sampling}
\label{sec:TS}

% Thompson sampling formula
Thompson Sampling, also known as randomized probability matching, has been empirically proven with satisfactory performance~\cite{chapelle2011empirical} and provable optimality properties with theoretical guaranteed regret bounds~\cite{agrawal2013further,russo2016information}.
Given the observed past history $D$ where $D$ is composed of triplet ($x_{1:t},a_{1:t},r_{1:t}$) and some prior distribution $P(\theta)$, the posterior distribution is given by Bayes rules, $P(\theta|D) \propto \Pi_{t-1}^{T} P(r_t|a_t,x_t,\theta)$.
Probability matching heuristic consists of randomly
selecting an action $a$ according to its probability of being optimal instead of choosing the action that maximize the immediate expected reward.
The probability will be marginalized over the posterior probability distribution of the parameters after observed data $D$ as follow,
\begin{equation}
    P[a=a_t^*|D] = \int \mathbb{I}[E(r|a,x,\theta) = max_a E(r|a,x,\theta)] P(\theta|D)\mathrm{d}\theta
\label{eq:ts-proba}
\end{equation}
.

While TS can solve the polytope arm set case in polynomial time~\cite{dani2008stochastic}, objective function like Equation~\ref{eq:argmax} cannot be solved since it is an NP-hard problem~\cite{agrawal2013further}.

% advantage: maybe move to intro
% 1. robust to delayed update
% 2. automatic handle uncertainty by network
% 3. difficult to find ucb argmax, but posterior sampling
% Thompson Sampling has several particularly appealing properties for our problem setting among other alternative algorithms (i.e. UCB).
% Thompson Sampling is more robust than UCB when the reward delay is long~\cite{chapelle2011empirical}.
% Moreover, UCB-like algorithms could not scale with a large number of arms $N$ . 
% It has to keep track of the upper bound confidence for each arm with an $NxN$ matrix.
% Moreover, finding the maximum of the upper confidence index in every time step is non-trivial since it is multimodal.
% It is generally assumed that evaluating reward function $f$ is more costly than maximizing the UCB index~\cite{srinivas2009gaussian}.
% Thompson Sampling can be performed with one single neural network.

% \subsection{Approximate Bayesian Inference}
% approximate methods
In order to perform Thompson Sampling algorithm with neural network, we apply approximate Bayesian inference method.
Popular approximate sampling methods include Markov Chain Monte Carlo (MCMC)~\cite{andrieu2003introduction}, Stochastic Gradient Descent~\cite{mandt2017stochastic}, Variational Inference (VI)~\cite{blei2017variational} and Dropout~\cite{gal2016dropout}.
% , Sequential Monte Carlo~\cite{doucet2009tutorial}
Here we adopt Concrete Dropout~\cite{gal2017concrete} which is a data-driven approximate inference with good performance and calibrated uncertainties that can be directly performed end-to-end on neural networks.
The illustration of the method is in Figure~\ref{fig:TS}.
First, we learn a value function through Maximum Likelihood Estimation (MLE) with an estimation model (neural network) to predict quality for each arm.
% as shown in Equation~\ref{eq:value}
% \begin{equation}
%     P(r=1|x,a,\theta) = \sigma(\theta(x,a))
% \label{eq:value}
% \end{equation}
The estimation model $\theta$ is trained regularly with either regression or classification loss depending on the task with observed history triplet data $D$.
The objective to optimize the binary cross-entropy loss is shown in Equation~\ref{eq:value-obj}, where the reward is a binary variable and $\theta$ is the model parameter.
\begin{equation}
    \theta^* = \argmax_\theta \sum_{i=1}^T r_i ln(\theta(x_i,a_i)) + (1-r_i)ln(1-\theta(x,a))
\label{eq:value-obj}
\end{equation}
.

Next, we apply Concrete Dropout as approximate Bayesian inference in our estimation model and the inference of our trained neural network with dropout activated will act as posterior sampling from the approximated distributions.
More precisely, each action will be chosen according to Equation~\ref{eq:ts-proba} where the value function is a posterior distribution and each model inference will be posterior sampling.
%Neural network with approximate Bayesian inference can then perform Thompson Sampling. 
As described in Section~\ref{sec:TS}, TS algorithm selects each arm based on its probability of being optimal with given context.
However, the true posterior distribution in Equation~\ref{eq:ts-proba} is intractable.
Instead of computing the integral in Equation~\ref{eq:ts-proba}, we draw a random parameter sample from the posterior, and select the arm that maximizes the expected reward.
That is,
\begin{equation}
\label{eq:argmax-ts}
    a_t^* = \argmax_a \theta_{t}(x_t), \text{  where  } \theta_t \sim f(\theta| D_{1:t-1})
\end{equation}
.

To draw a random parameter sample $\theta_t$ from the posterior, we draw one set of random masks as weights in the dropout layers and fix the mask throughout entire time step $t$.
To this end, Equation~\ref{eq:argmax} in TS that selects the optimal arm becomes Equation~\ref{eq:value-arg} with random drawn dropout masks.
\begin{equation}
    a^*_t = \argmax_{a\in A} P(r=1|x,a,\theta_t)
    \label{eq:value-arg}
\end{equation}

\subsection{Upper Confidence Bound Algorithm}
% NEWADD
Upper Confidence Bound (UCB) algorithm is a well known algorithm that follows
the principle of optimism in the face of uncertainty to apply efficient exploration.
There is a line of extensive work on UCB algorithms for both linear and nonlinear cases~\cite{chu2011contextual,srinivas2009gaussian,valko2013finite}.
The UCB algorithm consists of estimated reward and uncertainty whose action is selected to maximize the upper confidence bound:
\begin{equation}
    a^{*}_t = \argmax_{a\in A} \hat{r}_{a,t} + \hat{c}_{a,t}
\end{equation}
.

\begin{figure}[t]
    \centering
    \begin{subfigure}{0.45\textwidth}
        \includegraphics[width=\linewidth]{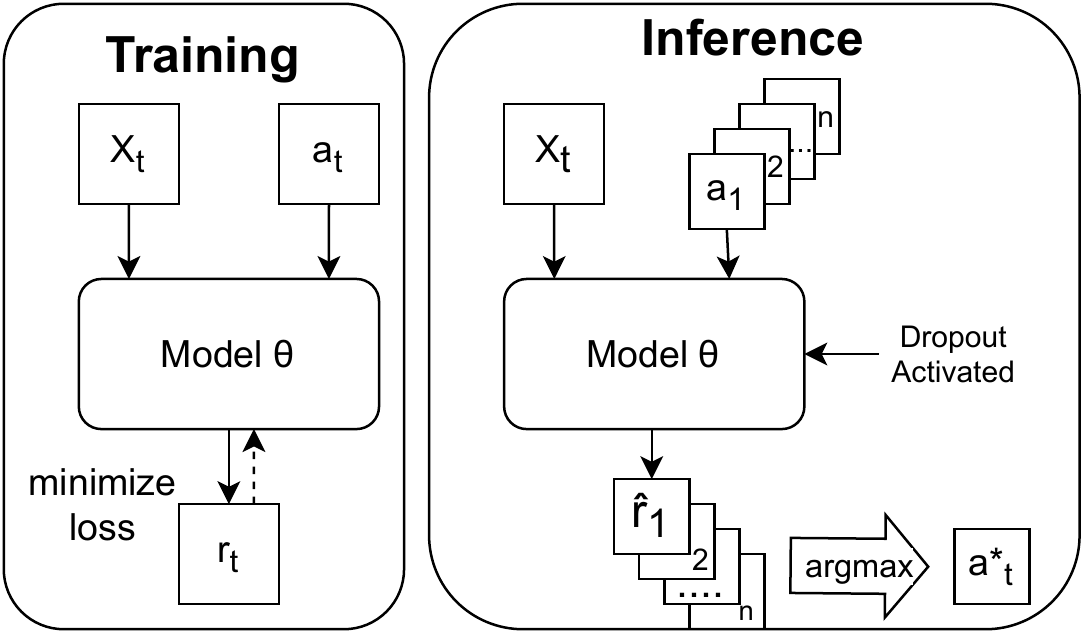}
        \caption{Performing Thompson Sampling approximate Bayesian inference.}
        \label{fig:TS}
    \end{subfigure}\hfill
    \begin{subfigure}{0.45\textwidth}
        \includegraphics[width=\textwidth]{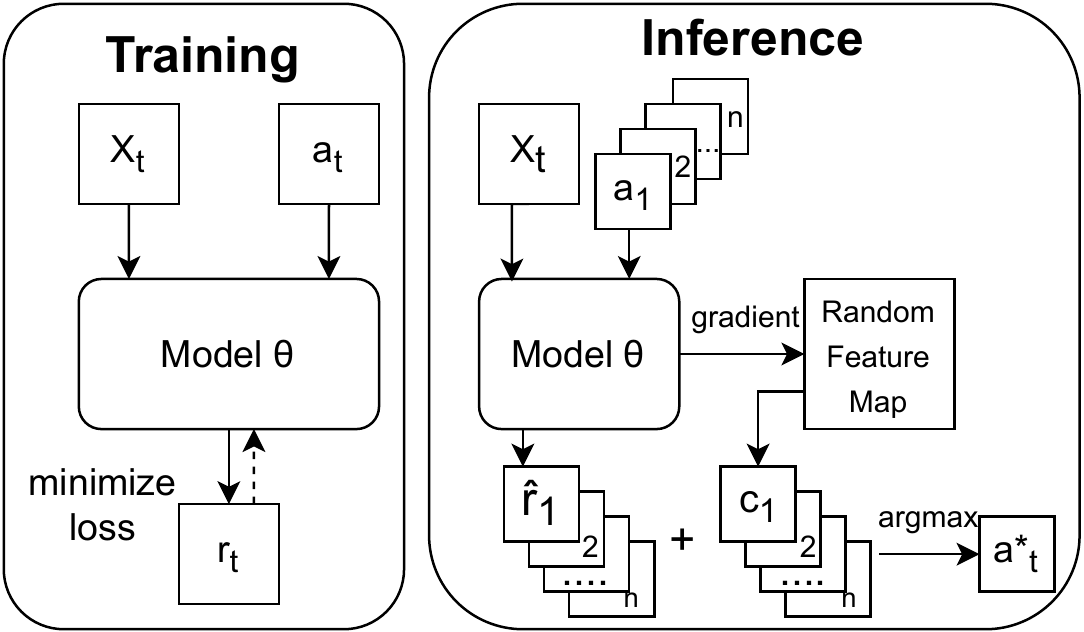}
        \caption{Performing NeuralUCB with gradient random feature mapping.}
        \label{fig:UCB}
    \end{subfigure}
    \caption{Basic neural bandit models}
    \label{fig:regular}
\end{figure}

Here we build on top of NerualUCB~\cite{zhou2020neural}.
The key idea of NeuralUCB  is to use a neural network $f(x; \theta)$ to predict the reward of context $x$, and upper confidence bounds computed from the network to
guide exploration~\cite{auer2002nonstochastic} through random feature mapping defined by the neural network gradient.
NeuralUCB has appealing properties that utilize the expressive power of deep neural networks with no assumption made about the reward function and has a differentiable objective function.
The upper confidence bound is computed by the following formula:
\begin{equation}
    U_{t,a} = f(x_{t,a}; \theta_{t-1})  + \gamma_{t-1}\sqrt{g(x_{t,a}; \theta_{t-1})^\top Z_{t-1}^{-1}g(x_{t,a}; \theta_{t-1})/m}\\
    % a_t =\argmax U_{t,a}
    \label{eq:ucb-obj}
\end{equation}
,

where $f$ is the neural network, $Z$ is the covariance matrix, $\gamma$ is the confidence scaling factor, $g(x; \theta)$ is the gradient $\nabla_{\theta} f(x; \theta) \in \mathbb{R}^p$ and $m$ is the network width.
Figure~\ref{fig:UCB} illustrates the method.

\section{Inference Scalability}
\subsection{Problem Definition}
\label{sec:definition}
% large scale, delayed, pre-compute
For online real-time inference, time complexity linear to the number of arms is usually not acceptable when the number of arms grows.
%Here we mainly want to address such scalability issues for time and space (memory) complexity with a large number of arms.
%In our problem setting, we aim at large scale real world system which the rewards is typically not processed immediately because of various run-time constraints.
We assume the system performs batch reward updates periodically as does in many existing works~\cite{chapelle2011empirical,zhou2020neural,zahavy2019deep}. As shown in Figure~\ref{fig:regular}, the input data utilize the reward of certain action to train a model estimating the reward.
During inference, each action together with the context vector is fed iteratively as the input to the model to obtain one action with the highest reward, as shown in Equation~\ref{eq:value-arg}. That means, each time step requires iterating through all $N$ arms, which can be very slow with large $N$. We will describe a more scalable approach to alleviate such a burden.

\begin{figure}[t]
    \centering
    \begin{subfigure}{0.46\textwidth}
        \includegraphics[width=\textwidth]{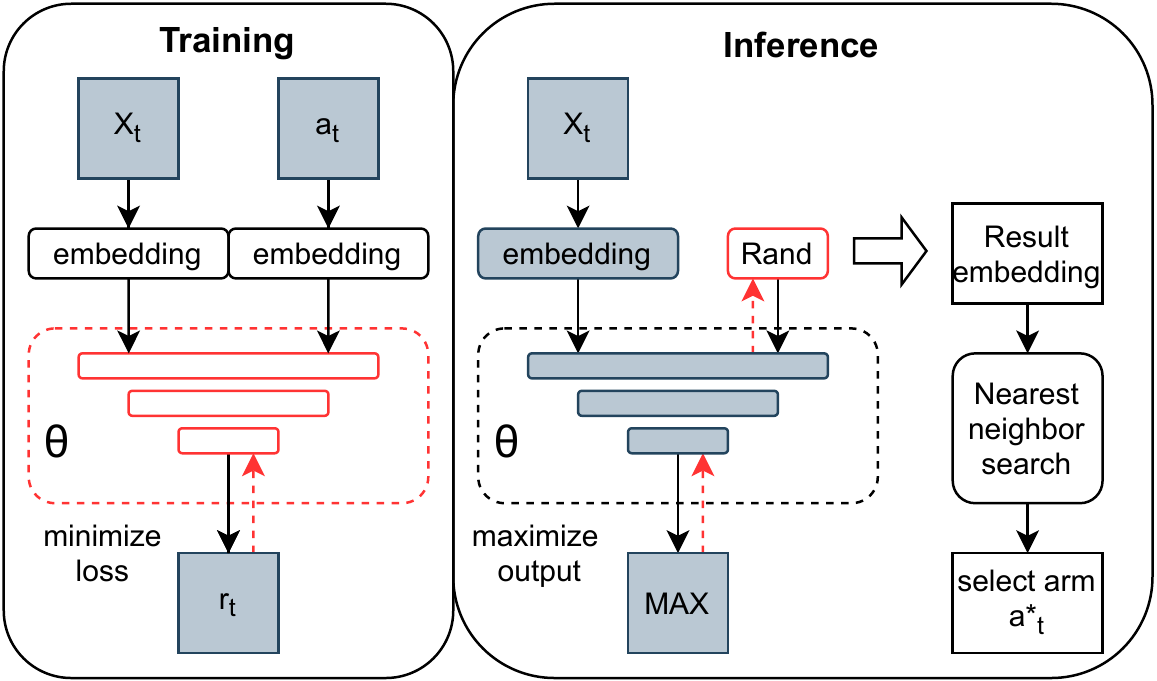}
        \caption{FastBandit Inference with Thompson Sampling.}
    \end{subfigure}\hfill
    \begin{subfigure}{0.53\textwidth}
        \includegraphics[width=\textwidth]{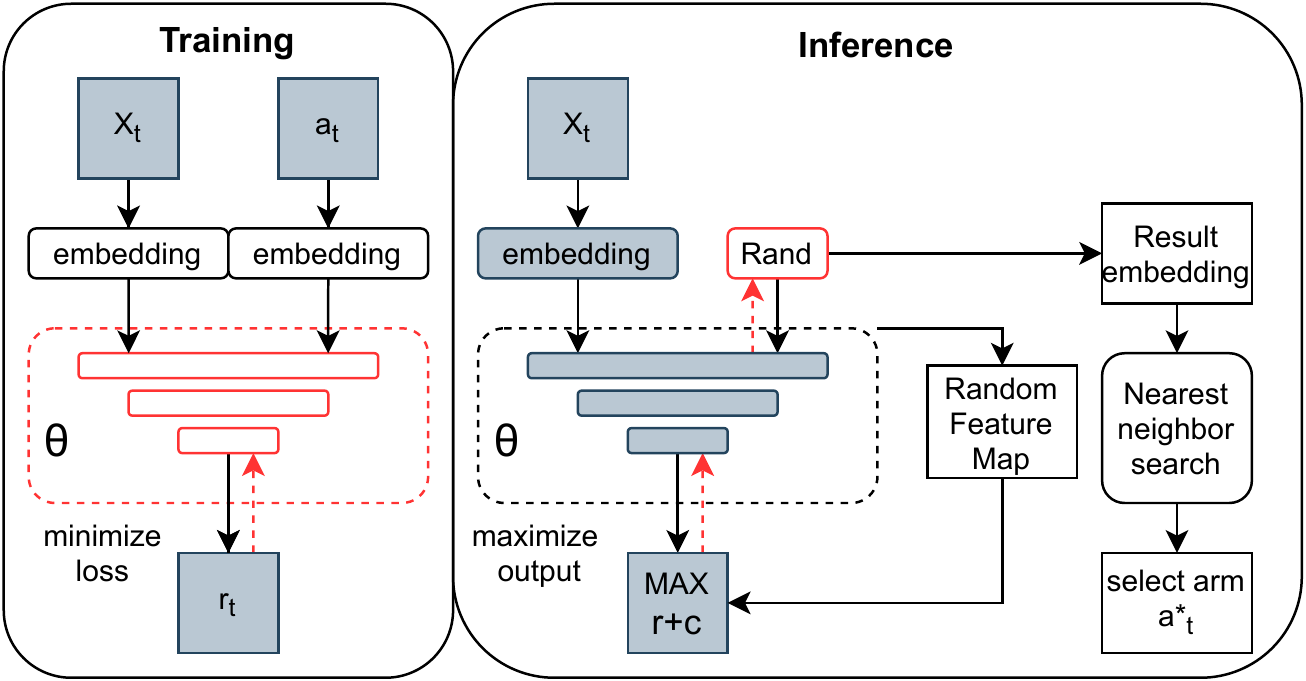}
        \caption{FastBandit Inference with UCB algorithm.}
    \end{subfigure}
    \caption{The red dashed lines represent the gradient passing and the red boxes highlight the optimization targets. The solid gray blocks indicate fixed or given attributes. The input embedding is sampled uniformly and gradient ascent is run to maximize output estimation value. Then ANNS is performed to search for the optimal arm.}
    \label{fig:backprop}
\end{figure}

\subsection{Inference the Best Arm in Logarithmic Complexity}
\label{sec:msga}

% Multistart stochastic gradient ascent~\cite{marti2003multi,marti2013multi}, which is a heuristic search procedures that is commonly used in Bayesian optimization, such as Knowledge gradient acquisition function that has a similar optimization problem to ours that runs on Gaussian process~\cite{wu2016parallel}. selects the best local optimum found as an approximate global optimum.
% Multistart methods is commonly used in Bayesian optimization, such as Knowledge gradient acquisition function which has a similar optimization problem to ours that runs on Gaussian process~\cite{wu2016parallel}.

%Neural networks with approximate Bayesian inference has a strong advantage that gradients can be calculated and pass through the network easily.
%Inspired by a optimization task in knowledge gradient acquisition function (Bayesian optimization) that runs on Gaussian process~\cite{wu2016parallel} using a heuristic search procedures for finding approximate global optimum based on Multistart methods~\cite{marti2003multi,marti2013multi}, we proposed the fixed-weight Back-Propagation (FastBandit) for efficiently identifying the best arm.
%The illustration of the process is in Figure~\ref{fig:backprop}.

First, we request that all input context features and arms be mapped into the embedding space during training, as shown in Figure~\ref{fig:backprop}, such that we can perform efficient gradient methods with the neural network.
At inference phase, an instance feature $x_t$ plus a random vector (as initial values for action embedding) are fed into a well-trained neural network $\theta$. We then generate the gradient based on the loss between predicted and optimal values. Such gradient is back-propagated to update the action embedding (i.e. marked as the red rectangle box $Rand$). We call such a process fixed-weight back-propagation because, instead of using back-propagation technique to update the weights of the model, here we use it to infer the embedding that can lead to the target optimal value, with the model weight remaining fixed. 

Inspired by the optimization task in Knowledge Gradient acquisition function (Bayesian optimization) that runs on Gaussian process~\cite{wu2016parallel} using a heuristic search procedure to find approximate global optimum based on Multistart methods~\cite{marti2003multi,marti2013multi}, here we adopt the multi-start methods to perform multiple instances of stochastic gradient ascent ~\cite{robbins1951stochastic,blum1954multidimensional} from different starting points and selects the best local optimum found as an approximate global optimum.

After multiple runs of gradient ascent, we acquire the action embeddings $\in R^d$ maximizing the estimated rewards.To map such embedding to an existing arm, we adopt approximate nearest neighbor search (ANNS) to find the arm with the nearest embeddings. The search time for $N$ elements in high dimension space scales with logarithmic complexity~\cite{malkov2018efficient}.
Recent advances in ANNS provide highly optimized software with distributed search~\cite{malkov2012scalable} and vector quantization~\cite{guo2020accelerating}, which is generally much faster than the gradient ascent step stated before. % beaumont2007peer
The overhead of quasilinear construction time for ANNS can be ignored since using batch update can ease the construction time through pre-computing.
The time complexity of one single arm selection is $O(log(N) + C)$ where $C=I\cdot R$ is constant number of iterations for multi-start stochastic gradient ascent.
We call this solution the FastBandit Inference.
Detailed steps are described in Algorithm~\ref{alg:msga}.

\subsection{Algorithm}
\label{app:algo}
\begin{algorithm}
\caption{FastBandit method for finding $x_t$ in Equation~\ref{eq:argmax-ts}, based on multistart stochastic gradient ascent and approximate similarity search.}
\label{alg:msga}
{\fontsize{8}{8}\selectfont
\begin{algorithmic}[1]
\Require ~~\\
~$R$: The number of runs, $I$: Iterations for each run of stochastic gradient ascent, $s$: The parameter used to define step size, $\tau$: The threshold for stop criterion\\

\State Set $Maxima = 0$, Generate $\theta_t \sim f(\theta|\;D_{1:t-1})$
\For{$r=1$ to $R$}
    \State Choose $x^r_0$ uniformly at random from metric space of $A$.
    \For{$i=1$ to $I$}
        \State Let $G$ be the gradient estimate of $\nabla\theta_t(x^r_{i-1})$
        \State Let $\alpha_t = s/(s+i)$
        \State $x^r_i = x^r_{i-1} + \alpha_t\cdot G$
        \IIf{$\theta_t(x^r_i) > \tau$} break \EndIIf
    \EndFor
    \State $x^r_I\leftarrow$ = Find Nearest Neighbor $x^r_I$
    \State $Maxima\leftarrow max(\theta_t(x^r_i),\;Maxima)$
\EndFor
\Return Maxima
\end{algorithmic}
}
% \vspace{-0.1cm}
\end{algorithm}

\begin{algorithm}
\caption{Minibatch stochastic gradient descent training of generative adversarial networks.}
\label{alg:gan}
\begin{algorithmic}[1]
\Require ~~\\
$k_d,k_g$: The number of steps applied to $D, G$\\

\For{Training iterations}
    \For{$k_d$ steps}
        \State $\bullet$Sample minibatch $\{(x_1,a_1,r_1), \cdots, (x_m,a_m,r_m)\}$
        \State{$\bullet$Update the discriminator by descending gradient:
        % \[
            % [\nabla_{\theta_d} \frac{1}{m} \sum_{i=1}^m \left[
            % r_i\cdot\log D\left(x_i,a_i)\right)
            % \right]]
            }
        % \]}
    \EndFor
    \For{$k_g$ steps}
        \State$\bullet$Sample minibatch of noize $\{z_1, \cdots, z_m\}$ from prior $p_z$
        \State$\bullet$Draw $\theta_{d}^\prime \sim f(\theta_d|M)$
        \State{$\bullet$Update the generator by ascending gradient:
        % \[
            % [\nabla_{\theta_g} \frac{1}{m} \sum_{i=1}^m
            % \log \left(1-D\left(G\left(z_i,x_i\right),\;\theta_{d}^\prime\right)\right)]
            }
        % \]}
    \EndFor
\EndFor
\end{algorithmic}
% \vspace{-0.1cm}
\end{algorithm}

\subsection{GANBandit: Shifting Computation from Inference to Training}
\label{sec:ganbandit}
In the previous section, we improve the time complexity for a single arm selection to $O(log(N) + C)$.
However, this solution is not without concerns.
First, it is known that in deep neural networks back-propagation is significantly slower than forward passing.
This implies that the optimization process of applying back-propagation for gradient ascend is much slower than computing the estimated value of a single arm. 
Second, given the previously proposed approach, the batch process requires copying the entire computation graph (or model) for each run in order to process multiple runs in parallel, which imposes a serious burden in terms of space complexity.
%In contrast, we can forward a large number of arms in one batch through a single computation graph simultaneously to acquire the output value from the estimation model.
Finally, back-propagation with larger $R$ or $I$, although enjoys a more accurate approximation, can result in longer latency C for real-time services.
There is a trade-off between minimizing approximation error and shortening service latency.

\begin{figure}[t]
    \centering
    \begin{subfigure}{0.49\textwidth}
        \includegraphics[width=\textwidth]{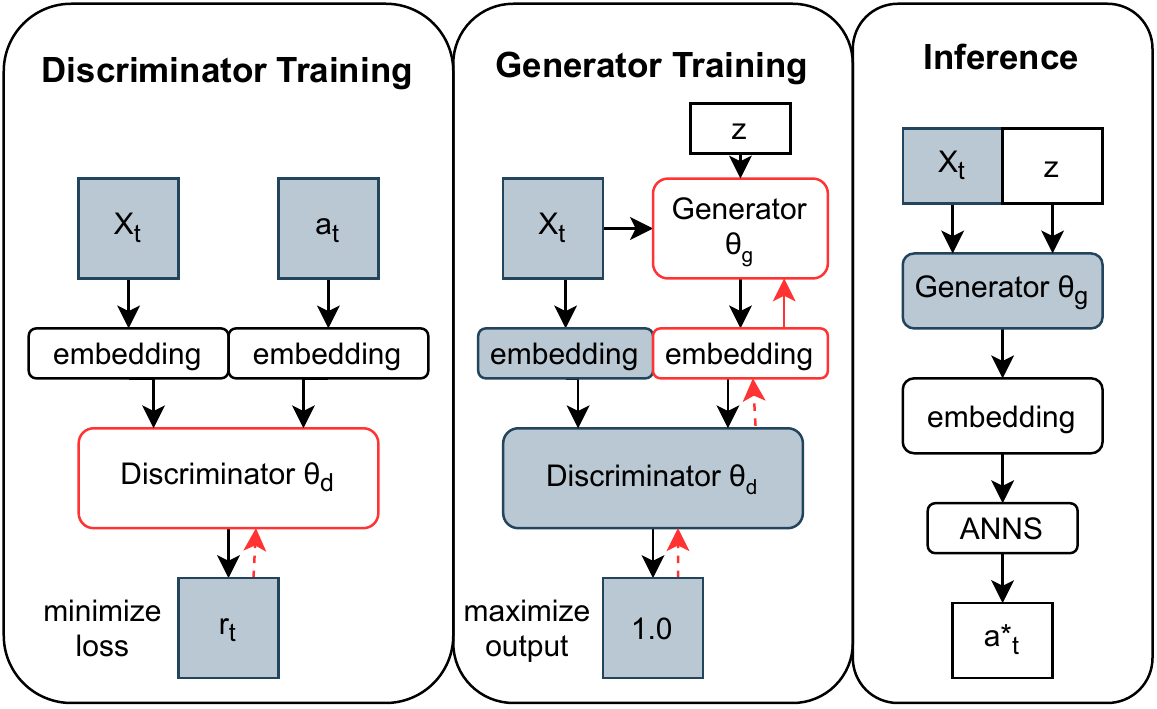}
        \caption{GANBandit for Thompson Sampling.}
    \end{subfigure}\hfill
    \begin{subfigure}{0.49\textwidth}
        \includegraphics[width=\textwidth]{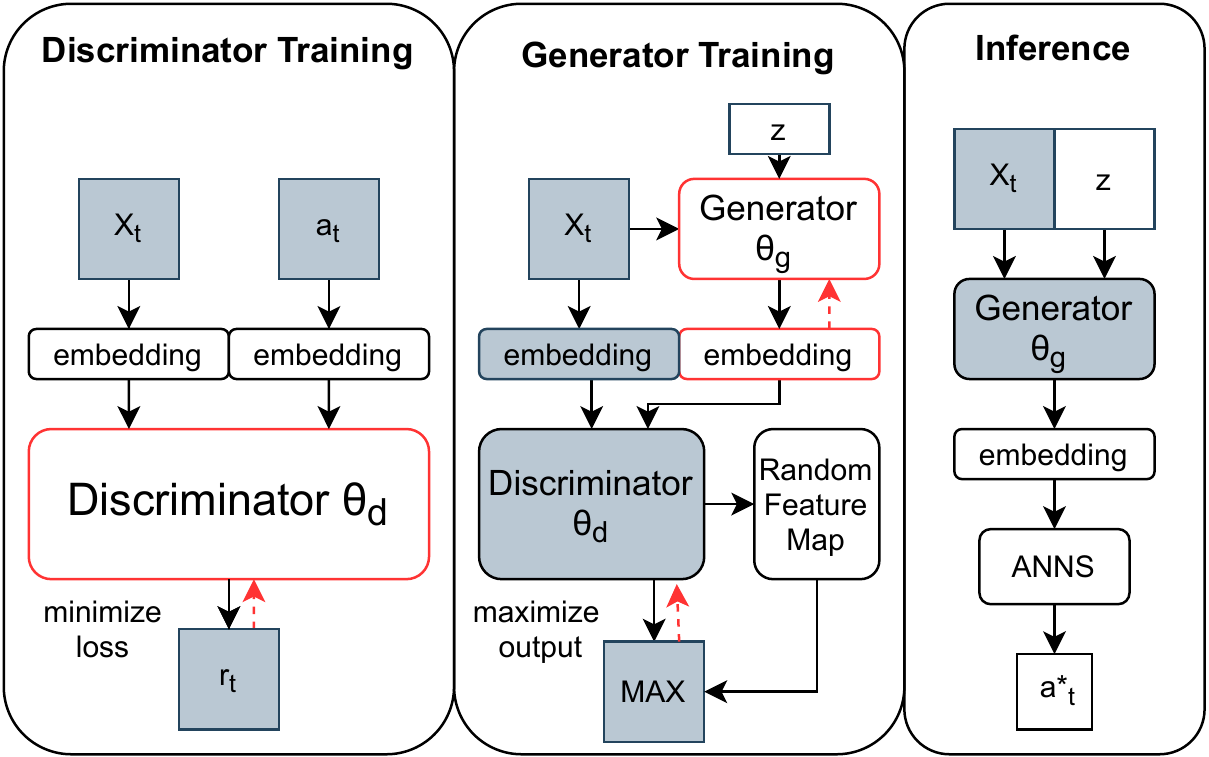}
        \caption{GANBandit for UCB algorithm.}
    \end{subfigure}
    \caption{The red dashed lines are the gradient passing and the red boxes are the optimized targets. The solid gray blocks indicate fixed or given attributes. The generator is jointly trained and tries to maximize the output value of the discriminator.}
\label{fig:gan}
\end{figure}

Here we propose a solution to move the optimization process of back-propagation gradient ascent from inference time to training time.
The main idea is to train a generator using adversarial training strategy similar to generative adversarial networks (GANs)~\cite{goodfellow2014generative} to optimize the same objective as stochastic gradient ascent.
The generator will be jointly trained with a reward estimation model (discriminator) at training stage such that the optimization of the gradient ascent no longer needs to be performed at inference time.
The proposed architecture is shown in Figure~\ref{fig:gan}.
This is a significant advantage for real time services such that forward passing of the neural network model can generate the optimal arm in Equation~\ref{eq:argmax-ts} and enjoy the significant speedup of batch processing.
This also implies that we do not need to sacrifice approximation accuracy and can employ a more exquisite and time consuming optimization strategy without run-time constraints.

To learn the generator’s distribution $p_g$ over the probability of arms being optimal given context $x_t$ as in Equation~\ref{eq:ts-proba}, we define a prior on input noise variables $p_z(z)$, then represent a mapping to data space as $G(z,x_t;\;\theta_g)$.
We then define the reward estimation model that outputs a single scalar as the discriminator $D(x_t;\;\theta_d)$ with binary cross entropy loss similar to Equation~\ref{eq:value-obj}.
We simultaneously train $G$ to minimize $1 - D(G(x_t, z))$ %which is similar to Equation~\ref{eq:value-obj} 
such that the output of the generator will be the optimal arm that maximizes the reward.
The generator will learn the argmax of certain $\theta_{d,t}$ that is drawn with fixed dropout parameters while the latent variable $z$ is drawn from prior $p_z(x)$.
Interestingly, the training of the discriminator that predicts the reward value can also be viewed as a binary classification task to distinguish if it is the optimal arm to generate maximum reward.
\begin{equation}
    \min_{D}\;\mathbb{E}_{a\sim p(a^*_t)}[\log(D(x_t,a,\theta_{d_t}))]
\label{eq:binary-classification}
\end{equation}
,

where $a^*_t$ is the true optimal arm (input embedding vector) that maximizes the output value of the discriminator $D$ similar to Equation~\ref{eq:argmax-ts}.
That is, $a^*_t$ will be the exact arm that we would acquire by performing gradient ascent with back-propagation through $D$ in FastBandit method.
Likewise, the objective of the generator to maximize $D(G(x_t, z))$ can also be interpreted as trying to output an arm that the discriminator cannot distinguish from the true optimal arm.
Such interpretation forms the min max objective function that is identical to the objective of GAN.
The overall objective is as follow:
\begin{multline}
    \min_{G} \max_{D} \; \mathbb{E}_{a \sim p(a^*_t)}[\log(D(x_t,a,\theta_d))]\; + \mathbb{E}_{z\sim p(z),\;\theta_{d,t} \sim f(\theta_d)}[\log(1 - D(x_t,G(x_t,z), \theta_{d,t}))]
\label{eq:minmax}
\end{multline}
.

Eventually as the generator converges and the discriminator believes it as the optimal arm, the generator will output arms with the probability similar to those found in the FastBandit method.
Equation~\ref{eq:minmax} mostly follows the original GAN objectives and training procedures.
The detailed algorithm is in Algorithm~\ref{alg:gan}.
In practice, Equation~\ref{eq:minmax} may not provide sufficient gradient for $G$ to learn well.
Early in learning, when $G$ is poor, $D$ can reject samples with high confidence causing $\log(1 - D(G(z)))$ to saturate.
Instead, we can train to maximize $\log D(G(z))$ which results in the
same fixed point of the dynamics of $G$ and $D$ but provides much stronger gradients.
We later attempt to modify the objective using dropout directly for sampling instead of relying on latent variable $z$.
However, this variable can still be utilized at inference time to control uncertainty and force exploration to solve the under-exploration problems caused by approximation error~\cite{phan2019thompson}.

To this end, online computation time complexity comparison is listed in Table~\ref{tab:complexity}.

\begin{table}[b!]
    \centering
    \begin{tabular}{lcccc}
    \toprule
        ~& QGLOC & Regular & FastBandit & GANBandit\\
    \midrule
        Memory & $d^2\cdot$ & $d\cdot B$ & d & d \\
        Inference & $N^\rho \log N$& $N/B$ & $R\cdot I+\log N$ & $\log N/B$\\
    \bottomrule
    \end{tabular}
    \caption{Time and space complexity comparison under big-O. $d$ is the maximum dimension layer. $N$ is the number of arms. $T$ is the number of time steps. $B$ is the batch size which is a large constant. The time complexity ignores $d$ and $T$ and only focus on $N$.}
    \label{tab:complexity}
    \vspace{-0.8cm}
\end{table}

\section{Experiment}
\label{sec:dataset}
We experiment on both synthetic and real-world data and focus on scenarios with large number of arms.
Experiment details with different parameters are listed in Appendix~\ref{app:finetune}.
We mainly compare with the following algorithms:
\begin{enumerate}[noitemsep]
    \item \textbf{Random} : random selection.
    \item \textbf{Overall best arm} : The single arm with highest reward among all data. A weak baseline without considering context.
    \item \textbf{LinearTS} : Linear Thompson Sampling algorithm.
    \item Exhaust TS : Neural Thompson Sampling with Exhaustive search for every $N$ arm in every time step. 
    \item \textbf{Exhaust UCB}: NeuralUCB with Exhaustive search for every $N$ arm in every time step.
    \item \textbf{FastBandit TS}: proposed method in Section~\ref{sec:msga} with Thompson sampling
    \item GAN TS: GANbandit in Section~\ref{sec:ganbandit} with Thompson sampling
    \item \textbf{FastBandit UCB}: proposed method in Section~\ref{sec:msga} with UCB algorithm
    \item \textbf{GAN UCB}: GANbandit in Section~\ref{sec:ganbandit} with UCB algorithm
\end{enumerate}

\subsection{Artificial Dataset}
We first generate synthetic data with context dimension $d = 4$, number of arms $N = 10000$ and number of rounds $T = 5000$.
The context vector $x\in R^d$ is randomly sampled from $N(0, I)$ and normalized to have unit norm.
We investigate the following three nonlinear functions:
\begin{align}
    h_1(x) = x\cos(x^T) a + 0.25(x^Ta) &~& h_2(x) = 10(x^Ta)^2 &~& h_3(x) = cos(3x^Ta)
\end{align}
,

where $a$ is randomly sampled from $N(0, I)$ and normalized to have unit norm.
For each function $h_i(\cdot)$, the reward at round $t$ for action $a$ is generated by $r_{t,a} = h_i(x_t,a) + \xi_t$, where $\xi_t$ is Gaussian noise independently drawn from $N(0, 1)$.

\subsection{Real-World Dataset}
For real-world data, we take two public classification datasets Celeba~\cite{liu2015faceattributes} and Bibtex~\cite{tsoumakas2008effective} along with three public recommendation datasets OpenBandit~\cite{saito2020large}, MovieLens~\cite{harper2015movielens} and The Movie Dataset \footnote{https://www.kaggle.com/rounakbanik/the-movies-dataset} on Kaggle.
Detailed information of the datasets is listed in Appendix~\ref{app:tab} Table~\ref{tab:dataset}.

For classification dataset, we follow the \textit{classification-to-contextual-bandit transform} in~\cite{dudik2011doubly} and optimize the classification problem with bandit algorithms in a fashion similar to Bayesian optimization.
In short, for each time step $t$, the bandit agent is given an instance of data, label pair $(x\in R^d, y\in R^{nclass})$.
The bandit agent will decide which class (arm) to explore/exploit depending on the given feature $x$ as context and later reveal the reward based on the ground truth.
The detail of the transformation is in Appendix~\ref{app:dataset}

% \begin{table}[t!]
%     \centering
%     % \tabcolsep=0.5cm
%     \begin{tabular}{lcccccc}
%     \toprule
%         ~& $N$ arms & $d$ dimension & $T$ instances\\
%     \midrule
%         $h1$ & 10000 & 4 & 5000\\
%         $h2$ & 10000 & 4 & 5000\\
%         $h3$ & 10000 & 4 & 5000\\
%         Bibtex & 160 & 1836 & 7395\\
%         Celeba & 10177 & 1*84*64 & 70838\\
%         MovieLens & 9724 & 16 & 6100\\
%         Openbandit & 81 & 40 & 10000\\
%         The Movie dataset & 14210 & 16 & 1000\\
%     \bottomrule
%     \end{tabular}
%     \caption{Details of the datasets.}
%     \label{tab:dataset}
%     \vspace{-0.8cm}
% \end{table}

\begin{figure}[t!]
    \centering
    \begin{subfigure}{0.49\textwidth}
        \includegraphics[width=\textwidth]{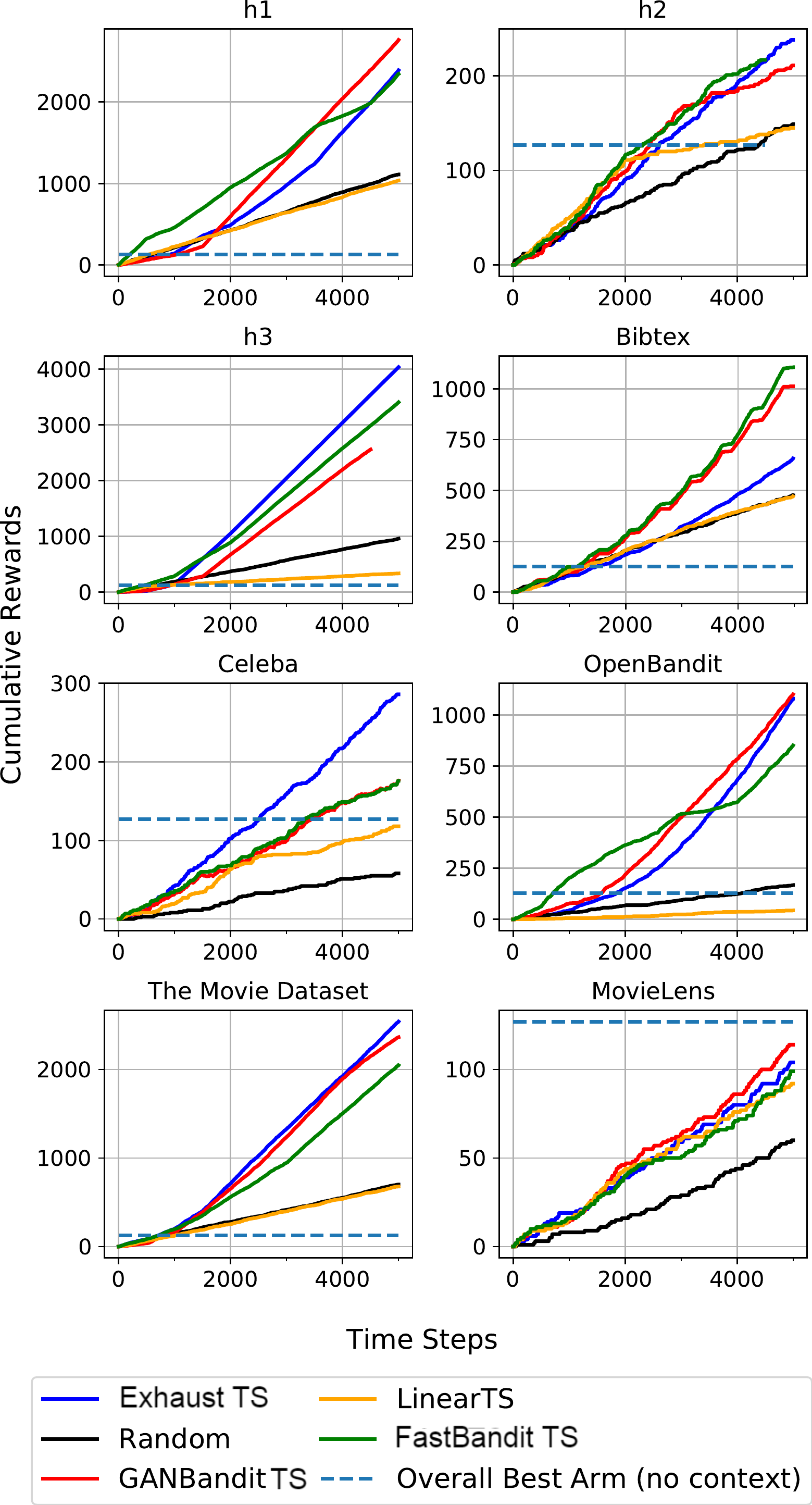}
    \end{subfigure}\hfill
    \begin{subfigure}{0.49\textwidth}
        \includegraphics[width=\textwidth]{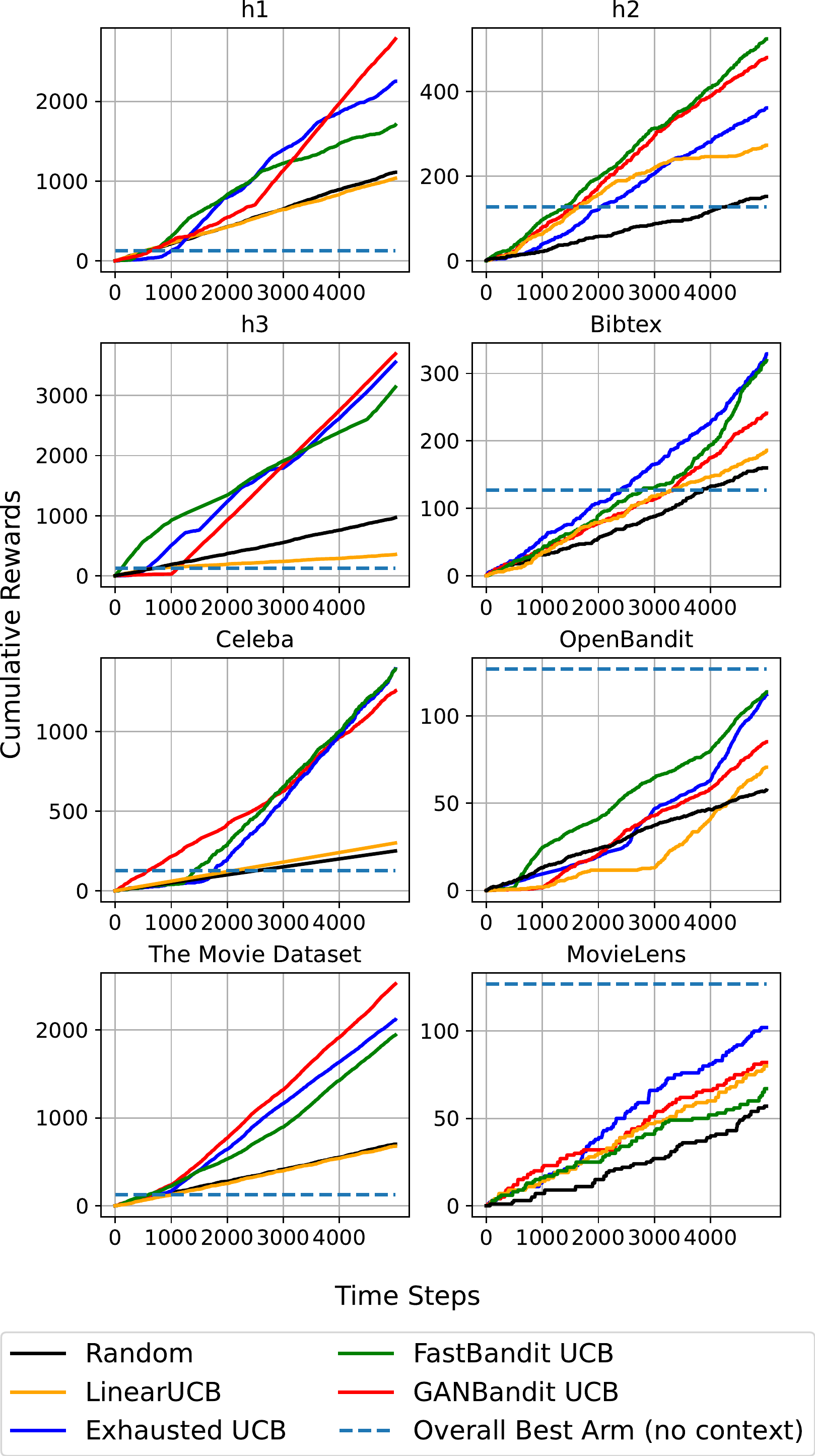}
    \end{subfigure}
    \caption{Comparison of the cumulative rewards.}
    \label{fig:regret}
\end{figure}

\subsection{Regret Bound Comparison}
The comparison of cumulative rewards of the 8 datasets is shown in Figure~\ref{fig:regret}.
First of all, we can observe that due to the nonlinearity of the reward function, LinearTS fail to learn the true reward function and hence results in almost linear regret for most dataset.
In contrast, by learning a more expressive representation and more efficient exploration, neural network models achieve sublinear regret which is much better.
Second, the cumulative rewards of two proposed approaches show competitive performance compared to the exhaustive search solutions.
This implies that the proposed algorithms capture nonlinearity of the underlying reward function. % while achieving a significant speedup simultaneously.
%This confirms the effectiveness of the proposed methods for contextual bandit problems with bounded nonlinear reward function.
Note that for each training instance, the arms are selected stochastically for querying such that the models are trained with different labels. Therefore, Exhaust search in our experiments does not necessarily  produce the best rewards.
%Noted that we did not show the results in the form of regrets since some of the dataset has not acquire enough time step to make the regret bounds obviously distinguishable between the baselines.

\subsection{Run Time Comparison}
\label{sec:runtime}
For the three models, Exhaust, FastBandit and GANBandit, that apparently outperform the others, we then compare their inference time. The run-time comparison for each algorithm is conducted on all 8 datasets.
We record the run-time for handling 100 sequential requests (given context $x_{1:100}$) during inference phase on a single GeForce RTX™ 3090 GPU.
The results are shown in Table~\ref{tab:runtime} while the unit is second per arm selection.

The run-time is measured in two different settings.
\textit{Single} measures the run-time of forwarding $N$ arms through the neural network one at a time.
This results in a much longer computation time compared to the batch process.
The results show that for single processing, \textit{FastBandit} is at least 3x faster than \textit{Exhaust} for six datasets with many arms.
On two datasets (Bibtex and Openbandit) with fewer arms, \textit{FastBandit} is not faster since the gain through back-propagation inference with limited number of arms cannot compensate the difference between forward and backward propagation.
Nevertheless, \textit{GANBandit} outperforms the others with order-of-magnitude in terms of speed.
%\textit{GANBandit} is lower than the linear time of full exhaustive search.
In Table~\ref{tab:runtime}, \textit{Batch} measures the run-time with batch processing that forwards the arms in a batch through the neural network.
Noted that this batch process is different from batch update mentioned in Appendix~\ref{app:setup}.
The formerevaluates all $N$ arms together in one single arm selection at \emph{inference} time while the latter refers to updating the model parameters with mini-batch during \emph{training} time.
The results show significant speedup for \textit{GANBandit} compared to others while \textit{FastBandit} has extensive overhead when run in batch due to high memory consumption and high cost to copy models across threads and processes.
In practice, GANBandit can accelerate even more since it handles each request with $\frac{1}{B}$ less memory compared to others.
The training time for GANBandit is roughly 3 times longer compared to others, which is reasonable in our application scenarios.

% In Figure~\ref{fig:no-batch}, the run-time is measured without performing batch process and the $N$ arms is forwarded through the neural network one at a time.
% This results in a much longer computation time compared to batch process.
% The results show that the time complexity of the proposed method \textit{Multi} and \textit{GANbandit} is lower than the linear time full exhaust search.

% Meanwhile in Figure~\ref{fig:batch}, the run-time is measured with batch processing that forwards all arms through the neural network at once.
% The results show significant speedup for \textit{GANBandit} compared to others while \textit{Multi} cannot be processed in batch due to short of memory to copy models.
% In practice, GANBandit can accelerate even more since it handles each requests with $\frac{1}{B}$ less memory.

\begin{table}[t!]
\begin{minipage}{0.48\linewidth}
    \centering
    \tabcolsep=0.1cm
    \begin{tabular}{llll}
    \toprule
        \textbf{TS} & \textbf{Single} & \textbf{Batch} & \textbf{Train}\\
    \midrule
        % Linear & $\num{1.54e-4}$ & $\num{6.01e-5}$\\
        Exhaust & $\num{4.03}\;s$ & $\num{0.012}\;s$ & $\num{9.12}\;s$\\
        FastBandit & $\num{1.312}\;s$ & $\num{0.691}\;s$ & $\textbf{8.76\;s}$\\
        GANBandit & \textbf{1.56e-4}\;s & \textbf{1.43e-6\;s} & $\num{31.49}\;s$\\
    \bottomrule
    \end{tabular}
\end{minipage}
\hfill
\begin{minipage}{0.48\linewidth}
    \centering
    \tabcolsep=0.1cm
    \begin{tabular}{llll}
        \toprule
            \textbf{UCB} & \textbf{Single} & \textbf{Batch} & \textbf{Train}\\
        \midrule
            % Linear & $\num{1.54e-4}$ & $\num{6.01e-5}$\\
            Exhaust & $\num{32.09}\;s$ & $\num{0.112}\;s$ & $\textbf{8.41\;s}$\\
            FastBandit & $\num{7.367}\;s$ & $\num{4.558}\;s$ & $\num{9.16}\;s$\\
            GANBandit & \textbf{3.01e-4}\;s & \textbf{7.92e-6\;s} & $\num{52.93}\;s$\\
        \bottomrule
    \end{tabular}
\end{minipage}
\\
\caption{The average run-time of 8 dataset, including the proposed methods and baseline. \textbf{Single} handles each arm separately while \textbf{Batch} inference all arms simultaneously. GANBandit significantly improves the run-time for nonlinear methods and can even outperform linear bandits. FastBandit method defeats Exhaust Search but has a much larger overhead in batch setting. The detailed run-time for separate dataset is in Appendix~\ref{tab:detail-runtime}.}
\label{tab:runtime}
\vspace{-0.8cm}
\end{table}

\section{Conclusion and Future Work}
This paper shows that a generative adversarial network based solution can be exploited to solve scalability issues for nonlinear bandit problems. Theoretically we advanced the time complexity from linear to logarithmic. 
The experiments have shown order-of-magnitude gain on efficiency with competitive performance in terms of rewards obtained. Furthermore, we also show that the proposed model can be extended to handle continuum-arm bandit setting in Appendix~\ref{app:continuum}.
%Our extensive experiment results on 8 datasets corroborate the proposed methods with significant run-time speedup and prove the theoretically guaranteed regret bounds holds.
Future work includes the theoretical analysis of GANBandit finite-time regret as well as applying the GANBandit to other applications such as network parameter tuning.

% \begin{ack}
% Use unnumbered first level headings for the acknowledgments. All acknowledgments
% More information about this disclosure can be found at: \url{https://neurips.cc/Conferences/2021/PaperInformation/FundingDisclosure}.

% Do {\bf not} include this section in the anonymized submission, only in the final paper. You can use the \texttt{ack} environment provided in the style file to autmoatically hide this section in the anonymized submission.
% \end{ack}

\appendix

\section{Appendix}

\subsection{Tables}
\label{app:tab}

\begin{table}[h!]
    \centering
    \tabcolsep=0.15cm
    \begin{tabular}{lcccccccccccccccc}
    \toprule
        ~& $h1$ & $h2$ & $h3$ & Bibtex & Celeba & MovieLens & Openbandit & Movie\\
    \midrule
        $N$ arms & 10000 & 10000 & 10000 & 160 & 10177 & 9724 & 81 & 14210\\
        $d$ dim & 4 & 4 & 4 & 1836 & 1*84*64 & 16 & 40 & 16\\
        $T$ instances & 5000 & 5000 & 5000 & 7395 & 70838 & 6100 & 10000 & 1000\\
    \bottomrule
    \end{tabular}
    \caption{Details of the datasets.}
    \label{tab:dataset}
    \vspace{-0.8cm}
\end{table}

\begin{table}[b!]
\begin{minipage}{0.48\linewidth}
    \centering
    \tabcolsep=0.15cm
    \begin{tabular}{lll}
    \toprule
        \multicolumn{3}{c}{Thompson Sampling}\\
    \bottomrule
    \toprule
        \textbf{h1} & \textbf{Single} & \textbf{Batch}\\
    \midrule
        % Linear & $\num{1.54e-4}$ & $\num{6.01e-5}$\\
        Exhaust & $\num{4.12}\;s$ & $\num{0.014}\;s$ \\
        FastBandit & $\num{1.287}\;s$ & $\num{0.741}\;s$\\
        GANBandit & \textbf{\num{1.81e-4}}\;s & \textbf{\num{1.64e-6}\;s}\\
    \bottomrule
    
    \toprule
        \textbf{h2} & \textbf{Single} & \textbf{Batch}\\
    \midrule
        % Linear & $\num{1.01e-4}$ & $\num{6.16e-5}$\\
        Exhaust & $\num{4.102}\;s$ & $\num{9.43e-3}\;s$ \\
        FastBandit & $\num{1.094}\;s$ & $\num{0.912}\;s$\\
        GANBandit & \textbf{\num{7.49e-4}\;s} & \textbf{\num{9.69e-7}\;s}\\
    \bottomrule
    
    \toprule
        \textbf{h3} & \textbf{Single} & \textbf{Batch}\\
    \midrule
        % Linear & $\num{9.96e-5}$ & $\num{1.1e-4}$\\
        Exhaust & $\num{3.37}\;s$ & $\num{9.53e-3}\;s$ \\
        FastBandit & $\num{1.095}\;s$ & $\num{0.923}\;s$\\
        GANBandit & \textbf{\num{7.6e-4}\;s} & \textbf{\num{1.13e-6}\;s}\\
    \bottomrule
    
    \toprule
        \textbf{Bibtex} & \textbf{Single} & \textbf{Batch}\\
    \midrule
        % Linear & $\num{1e-4}$ & $\num{6.7e-5}$\\
        Exhaust & $\num{0.05}\;s$ & $\num{6.6e-4}\;s$ \\
        FastBandit & $\num{0.141}\;s$ & $\num{0.798}\;s$\\
        GANBandit & \textbf{\num{6.2e-4}\;s} & \textbf{\num{9e-7}\;s}\\
    \bottomrule
    
    \toprule
        \textbf{Celeba} & \textbf{Single} & \textbf{Batch}\\
    \midrule
        % Linear & $\num{1.2e-4}$ & $\num{7.6e-5}$\\
        Exhaust & $\num{5.74}\;s$ & $\num{0.031}\;s$\\
        FastBandit & $\num{0.108}\;s$ & $\num{5.030}\;s$\\
        GANBandit & \textbf{\num{7.6e-4}\;s} & \textbf{\num{6.2e-6}\;s}\\
    \bottomrule
    
    \toprule
        \textbf{OpenBandit} & \textbf{Single} & \textbf{Batch}\\
    \midrule
        % Linear & $\num{1.1e-4}$ & $\num{6.3e-5}$\\
        Exhaust & $\num{0.034}\;s$ & $\num{6.9e-4}\;s$ \\
        FastBandit & $\num{1.032}\;s$ & $\num{0.721}\;s$\\
        GANBandit & \textbf{\num{6.6e-4}\;s} & \textbf{\num{8.8e-7}\;s}\\
    \bottomrule
    
    \toprule
        \textbf{MovieLens} & \textbf{Single} & \textbf{Batch}\\
    \midrule
        % Linear & $\num{7.1e-5}$ & $\num{6e-5}$\\
        Exhaust & $\num{4.06}\;s$ & $\num{9.3e-3}\;s$ \\
        FastBandit & $\num{1.351}\;s$ & $\num{1.044}\;s$\\
        GANBandit & \maxf{\num{7.7e-4}\;s} & \textbf{\num{5.4e-5}\;s}\\
    \bottomrule
    
    \toprule
        \textbf{Movie} & \textbf{Single} & \textbf{Batch}\\
    \midrule
        % Linear & $\num{1.2e-4}$ & $\num{1e-4}$\\
        Exhaust & $\num{6.07}\;s$ & $\num{0.013}\;s$\\
        FastBandit & $\num{1.344}\;s$ & $\num{1.15}\;s$\\
        GANBandit & \textbf{\num{8.5e-4}\;s} & \maxf{\num{1.5e-5}\;s}\\
    \bottomrule
    
    \end{tabular}
\end{minipage}
\hfill
\begin{minipage}{0.48\linewidth}
    \centering
    \tabcolsep=0.15cm
    \begin{tabular}{lll}
    \toprule
        \multicolumn{3}{c}{UCB}\\
    \bottomrule
    \toprule
        \textbf{h1} & \textbf{Single} & \textbf{Batch}\\
    \midrule
        % Linear & $\num{1.54e-4}$ & $\num{6.01e-5}$\\
        Exhaust & $\num{30.54}\;s$ & $\num{0.148}\;s$ \\
        FastBandit & $\num{7.587}\;s$ & $\num{4.415}\;s$\\
        GANBandit & \textbf{\num{3.79e-4}}\;s & \textbf{\num{8.11e-6}\;s}\\
    \bottomrule
    
    \toprule
        \textbf{h2} & \textbf{Single} & \textbf{Batch}\\
    \midrule
        % Linear & $\num{1.01e-4}$ & $\num{6.16e-5}$\\
        Exhaust & $\num{29.102}\;s$ & $\num{0.093}\;s$ \\
        FastBandit & $\num{8.094}\;s$ & $\num{3.922}\;s$\\
        GANBandit & \textbf{\num{3.78e-4}\;s} & \textbf{\num{1.01e-6}\;s}\\
    \bottomrule
    
    \toprule
        \textbf{h3} & \textbf{Single} & \textbf{Batch}\\
    \midrule
        % Linear & $\num{9.96e-5}$ & $\num{1.1e-4}$\\
        Exhaust & $\num{25.56}\;s$ & $\num{5.41e-2}\;s$ \\
        FastBandit & $\num{3.219}\;s$ & $\num{2.988}\;s$\\
        GANBandit & \textbf{\num{1.2e-3}\;s} & \textbf{\num{5.83e-6}\;s}\\
    \bottomrule
    
    \toprule
        \textbf{Bibtex} & \textbf{Single} & \textbf{Batch}\\
    \midrule
        % Linear & $\num{1e-4}$ & $\num{6.7e-5}$\\
        Exhaust & $\num{0.43}\;s$ & $\num{5.6e-3}\;s$ \\
        FastBandit & $\num{0.802}\;s$ & $\num{6.221}\;s$\\
        GANBandit & \textbf{\num{1.2e-3}\;s} & \textbf{\num{4e-6}\;s}\\
    \bottomrule
    
    \toprule
        \textbf{Celeba} & \textbf{Single} & \textbf{Batch}\\
    \midrule
        % Linear & $\num{1.2e-4}$ & $\num{7.6e-5}$\\
        Exhaust & $\num{41.77}\;s$ & $\num{0.321}\;s$\\
        FastBandit & $\num{0.699}\;s$ & $\num{35.4}\;s$\\
        GANBandit & \textbf{\num{1.6e-3}\;s} & \textbf{\num{2.2e-5}\;s}\\
    \bottomrule
    
    \toprule
        \textbf{OpenBandit} & \textbf{Single} & \textbf{Batch}\\
    \midrule
        % Linear & $\num{1.1e-4}$ & $\num{6.3e-5}$\\
        Exhaust & $\num{0.274}\;s$ & $\num{5.5e-3}\;s$ \\
        FastBandit & $\num{6.79}\;s$ & $\num{5.103}\;s$\\
        GANBandit & \textbf{\num{1.4e-3}\;s} & \textbf{\num{5.1e-6}\;s}\\
    \bottomrule
    
    \toprule
        \textbf{MovieLens} & \textbf{Single} & \textbf{Batch}\\
    \midrule
        % Linear & $\num{7.1e-5}$ & $\num{6e-5}$\\
        Exhaust & $\num{35.11}\;s$ & $\num{6.3e-2}\;s$ \\
        FastBandit & $\num{7.041}\;s$ & $\num{7.191}\;s$\\
        GANBandit & \maxf{\num{1.7e-3}\;s} & \textbf{\num{2.4e-4}\;s}\\
    \bottomrule
    
    \toprule
        \textbf{Movie} & \textbf{Single} & \textbf{Batch}\\
    \midrule
        % Linear & $\num{1.2e-4}$ & $\num{1e-4}$\\
        Exhaust & $\num{44.08}\;s$ & $\num{0.123}\;s$\\
        FastBandit & $\num{7.644}\;s$ & $\num{7.95}\;s$\\
        GANBandit & \textbf{\num{1.7e-3}\;s} & \maxf{\num{8.1e-5}\;s}\\
    \bottomrule
    
    \end{tabular}
\end{minipage}
\caption{The run-time comparison of the proposed methods and baseline. Single handles each arm separately while Batch processes all arms simultaneously. GANBandit significantly improves the run-time for nonlinear methods and can even outperform linear bandits. FastBandit method defeats Exhaust Search but has a much larger overhead in batch setting.}
\label{tab:detail-runtime}
\end{table}

\clearpage

\subsection{Dataset}
\label{app:dataset}
For classification dataset, we follow the \textit{classification-to-contextual-bandit transform} in~\cite{dudik2011doubly} to transform it to bandit dataset.
The idea is that the ground truth label (multi-class or multi-label) is not known for each observation, only whether the label chosen by the bandit agent for each observation is correct or not.
By such, the bandit agent can learn more by exploring classes (arms) for which it is less certain about or it can exploit more rewards if it is confident about predicting the correct label of a certain class.
Note that since we care about scenarios with large numbers of arms, we need to focus on classification datasets with many labels. 
The Celeba dataset is a multi-class celebrity image classification dataset where each celebrity belongs to its own class.
There are about 10K arms here.
We filter out celebrities with fewer images than 30 and then transform the original images into gray scale (one channel) and resize to $84*64$.
Experiments on Celeba dataset are the only ones that use convolutional neural networks.
The Bibtex dataset is a multi-label text classification dataset with BOW features, containing tags that people have assigned to different papers (the goal is to learn to suggest tags based on features from the papers), which is publicly available under the Extreme Classification Repository\footnote{http://manikvarma.org/downloads/XC/XMLRepository.html}.

For the three recommendation datasets, user features are used as context features and the items are used as arms.
Both user and item are transformed into embedding with dimension $d=8$.
The click/no-click labels in OpenBandit dataset are used as discrete binary rewards.
The rating labels in MovieLens and The Movie Dataset are transformed into binary rewards according to the probability of rating.
The transform rule is shown as follow, $P(rate) = rate*0.2, rate\in \{0.5, 1.0,\cdots 5.0 \}$.

\subsection{Algorithm}
\label{app:algo}
\begin{algorithm}
\caption{FastBandit method for finding $x_t$ in Equation~\ref{eq:argmax-ts}, based on multistart stochastic gradient ascent and approximate similarity search.}
\label{alg:msga}
{\fontsize{8}{8}\selectfont
\begin{algorithmic}[1]
\Require ~ $R$: The number of runs, $I$: Iterations for each run of stochastic gradient ascent, $s$: The parameter used to define step size, $\tau$: The threshold for stop criterion

\State Set $Maxima = 0$, Generate $\theta_t \sim f(\theta|\;D_{1:t-1})$
\For{$r=1$ to $R$}
    \State Choose $x^r_0$ uniformly at random from metric space of $A$.
    \For{$i=1$ to $I$}
        \State Let $G$ be the gradient estimate of $\nabla\theta_t(x^r_{i-1})$
        \State Let $\alpha_t = s/(s+i)$
        \State $x^r_i = x^r_{i-1} + \alpha_t\cdot G$
        \IIf{$\theta_t(x^r_i) > \tau$} break \EndIIf
    \EndFor
    \State $x^r_I\leftarrow$ = Find Nearest Neighbor $x^r_I$
    \State $Maxima\leftarrow max(\theta_t(x^r_i),\;Maxima)$
\EndFor
\Return Maxima
\end{algorithmic}
}
\end{algorithm}

\begin{algorithm}
\caption{Minibatch stochastic gradient descent training of generative adversarial networks.}
\label{alg:gan}
\begin{algorithmic}[1]
\Require ~~\\
$k_d,k_g$: The number of steps applied to $D, G$\\

\For{Training iterations}
    \For{$k_d$ steps}
        \State $\bullet$Sample minibatch $\{(x_1,a_1,r_1), \cdots, (x_m,a_m,r_m)\}$
        \State{$\bullet$Update the discriminator by descending gradient:
        % \[
            % [\nabla_{\theta_d} \frac{1}{m} \sum_{i=1}^m \left[
            % r_i\cdot\log D\left(x_i,a_i)\right)
            % \right]]
            }
        % \]}
    \EndFor
    \For{$k_g$ steps}
        \State$\bullet$Sample minibatch of noize $\{z_1, \cdots, z_m\}$ from prior $p_z$
        \State$\bullet$Draw $\theta_{d}^\prime \sim f(\theta_d|M)$
        \State{$\bullet$Update the generator by ascending gradient:
        % \[
            % [\nabla_{\theta_g} \frac{1}{m} \sum_{i=1}^m
            % \log \left(1-D\left(G\left(z_i,x_i\right),\;\theta_{d}^\prime\right)\right)]
            }
        % \]}
    \EndFor
\EndFor
\end{algorithmic}
\end{algorithm}

\subsection{Experiment Setup Details}
\label{app:setup}
All experiments run the contextual bandit problems with batch size $B = 500$ and the number of rounds $T = 5000$.
$B = 500$ indicates that 500 instances and corresponding rewards are updated at once in training phase.
The FastBandit model considers parameters iteration $I = 30$, runs $R = 10$, topk $K = 1$ while GANBandit uses parameters topk $K = 3$, $k_d = 1$, $k_g = 3$.
The neural networks in our experiments are mostly identical except for the image classification task on Celeba dataset.
It is a 3 layer fully connected network with hidden size = 8, embedding size = 8, dropout regularization = 1e-1 and Leaky-ReLU activation.
For image classification, we put two additional convolutional layers and max-polling layers on top of the network.
All experiments train the neural networks with Adam optimizer with learning rate = 1e-3, weight decay = 1e-5 and 1000 iterations in each time step.
The experiment code is given in this anonymous github repository\footnote{https://anonymous.4open.science/r/c4e4ff08-03b2-4455-87f0-133dd8c22353/}.

\subsection{Fine Tune Parameters}
\label{app:finetune}

The proposed approach integrates four optimization processes as its sub-components and essentially depends on the optimality of the sub-component optimizations for the theoretical regret bound to hold.
In order to achieve minimal approximation error, hyper-parameter fine-tuning is necessary.
From our experience, it is best to fine tune the four optimization process in the following order:
\begin{enumerate}[noitemsep]
    \item Train a DNN as the estimated reward function.
    \item Use dropout as approximate Bayesian inference for posterior sampling.
    \item Use heuristic multi-start methods to approximate global maxima.
    \item Approximate nearest neighbor search in high dimension space.
\end{enumerate}
Beyond doubt, reward estimation is the most fundamental in the bandit algorithm and thus learning a good approximate reward function is a critical first step.
Then it becomes a regular fine-tune task for training neural networks.
After having a well-trained neural network, the next step is to fine tune approximate Bayesian inference with enough uncertainty that best suits the data.
In our experiments, We apply data-driven Concrete Dropout instead of regular dropout mechanisms such that a grid search for dropout rate is avoided, substituted by a dropout regularization parameter.
Next, fine tune the number of runs, iterations and learning rate for back-propagation to achieve well approximated global maxima for arm embeddings.
Finally, with the parameters obtained, we train the generator with adversarial training strategy and fine tune the parameters for well-distributed arm sample quality.
This final step is similar to training a regular generative adversarial network for which many stabilization training tricks can be applied~\cite{chintala2016train}.
% Comparisons of different parameters for the four approximation methods is in supplementary materials.

\subsection{Extension to Continuum-Arm Bandit}
\label{app:continuum}

%In this paper, we consider the structured stochastic contextual bandit problem and assume the arm set of an infinite cardinality.
In the previous sections we focus on a finite but very large number of arms for large scale online bandit systems.
In this section, we will demonstrate that the proposed methods are also applicable for continuum-arm bandits in generic metric space of arbitrary structures.

\begin{figure}[h!]
    \begin{minipage}{0.49\linewidth}
        \centering
        \includegraphics[width=\textwidth]{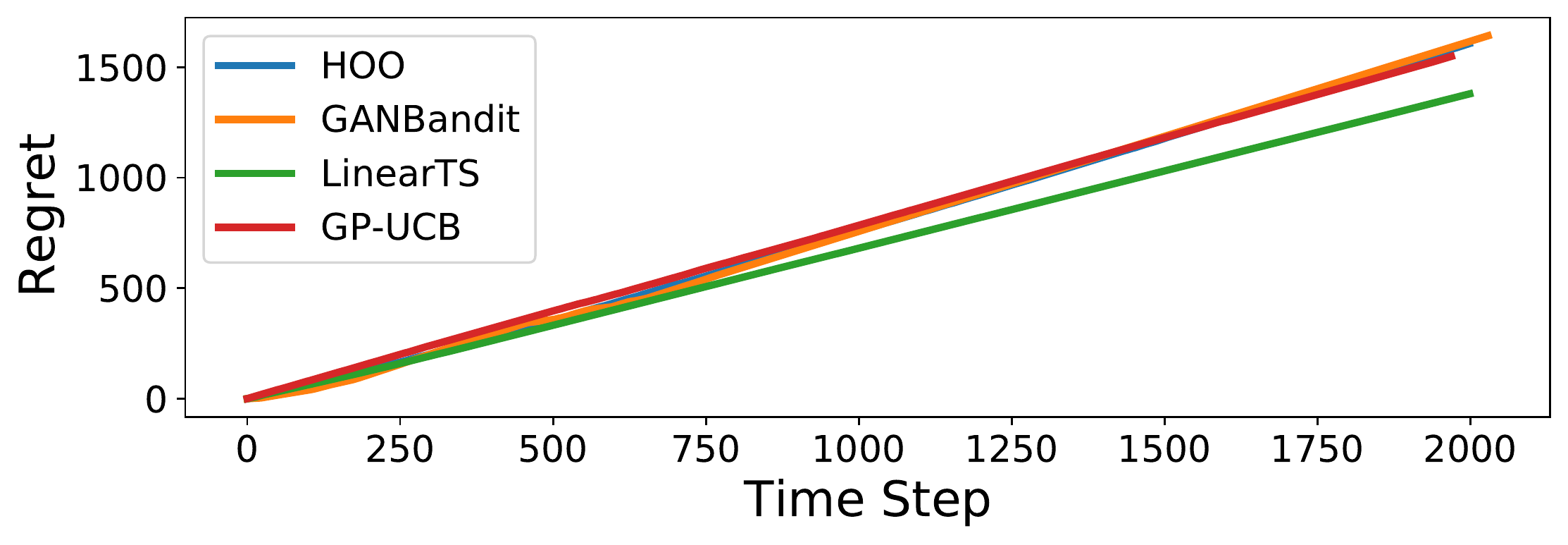}
        \caption{The cumulative reward comparison in continuum-arm setting.}
        \label{fig:inf-regret}
    \end{minipage}
    \hfill
    \begin{minipage}{0.48\linewidth}
        \centering
        \captionsetup{type=table}
        \begin{tabular}{lcccc}
        \toprule
            ~& GP-UCB & HOO & GANbandit\\
        \midrule
            Train & $O(1)$ & $O(T)$ & $O(T)$\\
            Inference & $O(T^3)$ & $O(T)$ & $O(1)$\\
        \bottomrule
        \end{tabular}
        \caption{The inference time complexity comparison between GP-UCB, HOO and GANBandit. Note that GANBandit requires a much longer training time.}
        \label{tab:con-complexity}
        \vspace{-0.8cm}
    \end{minipage}
\end{figure}

In a continuum-arm bandit there is no scalability issue caused by large number of arms since the number of arms are infinite.
However, the burden to maximize the objective of selecting the optimal arm for exploring/exploiting in Equation~\ref{eq:argmax} still exists.
Existing algorithms such as GP-UCB~\cite{srinivas2009gaussian}, GP-TS~\cite{chowdhury2017kernelized} that performs bandit algorithm on Gaussian processes have time and space complexity of at least quadratic to the number of dimensions $d$ and time step $T$.
Hierarchical optimistic optimization (HOO)~\cite{bubeck2011x}, one of the heuristic Monte-Carlo tree search methods with cumulative regrets as objective, also hasquadratic time complexity with time step $T$.
In contrast, the proposed method with approximate Bayesian inference with multi-start methods provides an efficient end-to-end algorithm that significantly reduce the time complexity to constant.
Moreover, GANBandit works exactly the same in continuum-arm bandit and achieve $O(1)$ time complexity during online inference.
Time complexity comparison for arm selection is shown in Table~\ref{tab:con-complexity}.

Modifications to partial back-propagation method and GANBandit for continuum-arm bandit are minimal.
The only change is that we do not need to perform nearest neighbor search since we do not need to discretize our acquired results to finite arms.
For the FastBandit model, the arm embedding vector after gradient ascent becomes our selected arm.
For GANBandit, the output embedding vector from the generator according to the probability of being optimal (maximize reward) becomes our selected arm.
%Originally, every optimization step in the proposed method are already processed through a neural network with an end-to-end fashion in metric space.
Similar to the experiment setup in Appendix~\ref{app:setup}, we again conduct experiments and compare with GP-UCB, HOO and LinearTS in regret bound.
LinearTS is performed by discretizing the metric space into 1000 arms.
The objective is to minimize cumulative regrets on the target function:
\begin{equation}
    h(x) = 0.5 * (np.sin(13 * x) * np.sin(27 * x) + 1),\;x\in[0, 1]
\end{equation}
% The regret bound is in Figure~\ref{fig:inf-regret} and the sample distribution is in Figure~\ref{fig:inf-dist}.
% These results demonstrates the effectiveness of GANBandit to perform Thompson sampling algorithm in continuum-arm bandit with well-distributed sample probability.
The result is shown in Figure~\ref{fig:inf-regret}.
The graph shows similar cumulative reward between GANBandit TS, GP-UCB and HOO algorithms while LinearTS clearly lags behind.
This shows that our model theoretically reduces the inference time from linear to constant and the approximation process does not apparently sacrifice the performance. 
%lower complexity the effectiveness of the proposed methods while being significantly faster at inference time.
% Thanks to the universal approximation theorems~\cite{csaji2001approximation}, deep neural network has the expressive power of fitting any arbitrary structures.

% \begin{figure}[t]
%     \centering
%     \includegraphics[width=0.5\textwidth]{Figs/inf-dist2.pdf}
%     \caption{The selected arm distribution between different algorithms.}
%     \label{fig:inf-dist}
% \end{figure}

\section*{References}

\bibliography{mybibfile}

\end{document}